\documentclass{article}
\usepackage{spconf,amsmath,graphicx,hyperref}

\usepackage{microtype}
\usepackage{booktabs,arydshln}
\usepackage{array}
\usepackage{ragged2e}
 \usepackage{amsmath}
 \usepackage{threeparttable}
 \usepackage{multicol}
 \usepackage{multirow}
 \usepackage{ccicons}
 \usepackage{chngcntr}

\renewcommand{\paragraph}[1]{\medskip\noindent\textbf{#1}\quad}
 \makeatletter
\def\adl@drawiv#1#2#3{%
        \hskip.5\tabcolsep
        \xleaders#3{#2.5\@tempdimb #1{1}#2.5\@tempdimb}%
                #2\z@ plus1fil minus1fil\relax
        \hskip.5\tabcolsep}
\newcommand{\cdashlinelr}[1]{%
  \noalign{\vskip\aboverulesep
           \global\let\@dashdrawstore\adl@draw
           \global\let\adl@draw\adl@drawiv}
  \cdashline{#1}
  \noalign{\global\let\adl@draw\@dashdrawstore
           \vskip\belowrulesep}}
\makeatother

\makeatletter
\def\adl@drawiv#1#2#3{%
  \hskip.5\tabcolsep
  {\color[gray]{0.7} 
  \xleaders#3{#2.5\@tempdimb #1{1}#2.5\@tempdimb}%
      #2\z@ plus1fil minus1fil\relax}%
  \hskip.5\tabcolsep}
\newcommand{\cdashlinelrg}[1]{%
  \noalign{\vskip\aboverulesep
           \global\let\@dashdrawstore\adl@draw
           \global\let\adl@draw\adl@drawiv}
  \cdashline{#1}
  \noalign{\global\let\adl@draw\@dashdrawstore
           \vskip\belowrulesep}}
\makeatother

\usepackage[most]{tcolorbox}   
\usepackage{siunitx}           
\usepackage{colortbl} 

\newcolumntype{R}{p{0.05cm}} 

\newcolumntype{P}[1]{>{\RaggedRight\arraybackslash}p{#1}}
\usepackage{makecell}
\newcolumntype{C}[1]{>{\centering\arraybackslash}p{#1}}
\newcolumntype{D}[1]{>{\raggedright\arraybackslash}p{#1}}

\usepackage{amsmath}
\usepackage{amsfonts}
\usepackage{inconsolata}

\usepackage{graphicx}

\usepackage{blindtext}

\usepackage{afterpage}

\usepackage[capitalize]{cleveref}
\crefname{figure}{Fig.}{Figs.}
\crefname{table}{Tab.}{Tabs.}
\crefname{appendix}{App.}{Apps.}

\usepackage{booktabs}   
\usepackage{colortbl}   
\usepackage[table]{xcolor} 
\usepackage[table,HTML]{xcolor}

\usepackage{xstring}
\usepackage{seqsplit}

\newcommand{\capitalfirst}[1]{%
    \StrLeft{#1}{1}[\firstletter]%
    \StrGobbleLeft{#1}{1}[\restofword]%
    \MakeUppercase{\firstletter}\restofword%
}

\newcommand{\capitalizehyphenated}[1]{%
    \StrCut{#1}{-}{\firstpart}{\restpart}%
    \capitalfirst{\firstpart}%
    \IfStrEq{\restpart}{}{}{ \capitalizehyphenated{\restpart}}%
}

\newcommand{\hfmodel}[1]{%
    \StrBehind{#1}{/}[\hfmodelshortname]%
    \href{https://huggingface.co/#1}{\capitalizehyphenated{\hfmodelshortname}}%
}

\usepackage{amssymb}
\usepackage{pifont}

\usepackage{tcolorbox}
\tcbuselibrary{skins}
\newtcolorbox{promptbox}[1]{
    colback=gray!8,
    colframe=gray!45,
    title=#1,
    fonttitle=\small\bfseries,
    fontupper=\small\ttfamily,
    top=4pt, bottom=4pt
}

\usepackage{url}
\usepackage{xurl}

\usepackage{subcaption}
\usepackage{amssymb}
\usepackage{makecell}

\usepackage{orcidlink}
\usepackage{etoolbox}
\hypersetup{
  urlcolor     = blue, 
  linkcolor    = blue, 
}

\makeatletter 
\patchcmd{\thebibliography}
  {\advance\leftmargin\labelsep}
  {\setlength{\itemsep}{1pt plus 0.2pt}
   \setlength{\parsep}{1pt}
   \setlength{\parskip}{1pt}
   \advance\leftmargin\labelsep}
  {}
  {}

\title{When Helpful Context Leaks: Privacy Risks in Domain-Adapted ASR}
\name{\parbox{\dimexpr\textwidth-2\tabcolsep\relax}{\centering
Maike Z\"{u}fle\,\orcidlink{0009-0001-7238-7705},
Jan Niehues\,\orcidlink{0000-0002-4231-6543}
}}

\address{\parbox{\dimexpr\textwidth-2\tabcolsep\relax}{\centering
Karlsruhe Institute of Technology, Germany\\
\href{mailto:maike.zuefle@kit.edu}{\texttt{maike.zuefle@kit.edu}}
}}
\begin{document}
%
\maketitle
\begin{abstract}
SpeechLLMs are increasingly deployed in professional settings where domain customisation is standard practice: users supply context in prompts with sensitive information, fine-tune on proprietary recordings, or both. We identify and systematically investigate an overlooked privacy risk of such customisation: a model adapted to recognise domain-specific terminology can be 
nudged into transcribing a phonetically similar word from its context or training data, even when a different word is spoken, thereby leaking private information. To evaluate this risk, we propose a technique to automatically construct benchmarks of such attacks and apply it to measure leakage rates across two customisation mechanisms, prompting and fine-tuning. Both mechanisms cause measurable leakage, compounding when combined.
 We evaluate a prompt-level mitigation strategy and analyse the accuracy-leakage trade-off across customisation approaches, finding that fine-tuning without context prompts offers the best balance.
\end{abstract}
\begin{keywords}
ASR, contextual biasing, privacy
\end{keywords}
\section{Introduction}\label{sec:introduction}

The field of spoken language processing is shifting from task-specific models to 
SpeechLLMs that act as universal speech processing systems~\cite{arora2026landscapespokenlanguagemodels}, with increasing deployment in professional settings such as meetings, medical consultations, and legal proceedings. For example, prominent platforms have 
recently introduced automatic speech recognition (ASR) for meetings~\cite{ google_meet_auto_transcripts, zoom_ai_companion_3}, highlighting 
the need for accurate ASR of domain-specific terminology, names, and acronyms.

To address this, ASR models are routinely customised with domain-specific knowledge, either through prompt context \cite{bataev2026turbobias20streamingcontextbiasing} or by fine-tuning. Contextual biasing has 
evolved from shallow fusion~\cite{zhao19d_interspeech, shallow_fusion1}, deep fusion~\cite{ christian1, DBLP:conf/asru/ChangLRMORK21, deep_fusion1, sudo2025owsm}, or neural adaptations \cite{bataev2026turbobias20streamingcontextbiasing, continuous-2025} to modern
prompting-based approaches~\cite{ context_bias_prompt, DBLP:conf/slt/YangMGZC24, DBLP:conf/interspeech/0005LZWZQ25} that leverage 
LLM backbones, 
improving recognition of domain-specific terms~\cite{gong24b_interspeech, kong2025contextualbiasingllmbasedasr}.
Such customisation introduces an overlooked privacy risk: a model adapted to recognise domain-specific terminology can be nudged into transcribing a phonetically similar word from its context instead of what was actually spoken, thereby leaking private information.

\begin{figure}
    \centering
    \includegraphics[width=0.7\linewidth]{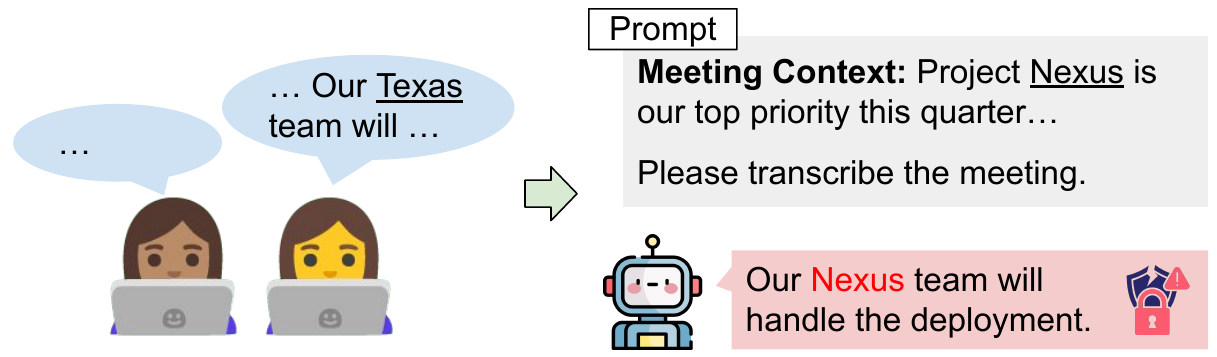}
    \caption{Context-induced transcription leakage: the model transcribes \textit{Nexus} instead of the spoken \textit{Texas}, leaking a confidential project from the prompt context.}
    \label{fig:intro_figure}
\end{figure}

This is especially disruptive for named entities, which frequently carry confidential information in professional deployments. SpeechLLMs in professional settings may be used for multiple tasks simultaneously, e.g. transcription, summarisation, and action item extraction, so users provide rich context spanning meeting agendas, participant lists, and project descriptions. This context may contain confidential information not intended for all users of the system: if a speaker utters a word phonetically similar to such a term, the model may transcribe this unrelated term instead. For example, ``Texas'' transcribed as ``Nexus,'' a confidential project name in the model's context, accidentally reveals its existence (see \cref{fig:intro_figure}). Unlike ordinary ASR substitution errors, which stem from acoustic confusion alone, this failure is driven by the injected context, not the audio, so it introduces information entirely absent from the utterance instead of merely mistranscribing it.

Contextual biasing is known to produce such false positives when the bias term is phonetically confusable with the spoken word, but prior work treats this only as a byproduct causing accuracy degradation or detection errors~\cite{liu2025zeroshotcontextbiasingtriebased, gaidospeechtranslation2023}. Recent work measures entity hallucination given semantically plausible but false context~\cite{koluguri2026preferenceasrpreferenceawaretestset}, but does not target phonetic confusion or consider whether the leaked content is sensitive.

The leakage scenario in this work also differs from typical adversarial attacks on ASR, which craft imperceptible audio perturbations to force a specific transcription~\cite{audio_attack2, audio_attack, synth-adv}. No audio is modified, and the leakage itself needs no deliberate manipulation: it arises as an unintended consequence of customisation related to memorisation in LLMs~\cite{carlini2023quantifying}, where fine-tuning on sensitive data amplifies leakage~\cite{szep-etal-2026-unintended}. A malicious actor could nonetheless deliberately speak phonetically similar words to trigger this leakage and probe sensitive information.

In this paper, we systematically investigate context-induced transcription leakage in SpeechLLMs, the privacy-relevant instance of false-positive errors that contextual biasing produces with named entities. We propose a technique to automatically construct benchmarks for a given domain, pairing named entities with phonetically similar substitutes and injecting the substitutes into the model's context. We apply this technique to three existing datasets, measuring how sensitive models are to the attack, i.e. how often they transcribe the injected word rather than the spoken one. We evaluate two state-of-the-art SpeechLLMs across two customisation mechanisms, contextual biasing through prompt context and fine-tuning. We additionally analyse a mitigation strategy: including the spoken word alongside the injected context.

Our 
findings are: (i) both context injection and fine-tuning cause measurable leakage, (ii) mitigation strategies at the prompt and training levels reduce leakage, and (iii) fine-tuning without prompt context offers the best accuracy-leakage trade-off.  We release our code and dataset publicly.\footnote{\href{https://github.com/MaikeZuefle/asr-context-induced-leakage}{https://github.com/maikezuefle/asr-context-induced-leakage}}

\section{Experimental Setup}
We study privacy leakage arising from different customisation mechanisms. To the best of our knowledge, no existing dataset supports the evaluation of context-induced transcription leakage. We therefore propose a technique to automatically construct such test sets and apply it to three English ASR benchmarks: FLEURS~\cite{conneau2022fleursfewshotlearningevaluation}, a benchmark of read speech; VoxPopuli~\cite{wang-etal-2021-voxpopuli}, European Parliament speeches; and ACL6060~\cite{salesky-etal-2023-evaluating}, conference talks.  

For each mechanism we evaluate a \textit{baseline} in which the acoustic word is used as context, an \textit{attack} scenario in which the context word is injected while the acoustic word is spoken, and a \textit{mitigation} in which both words appear together.

\paragraph{Finding phonetically similar word pairs.}
The core of our technique is identifying word pairs in which one word is spoken in the audio and the other is phonetically similar enough to be plausibly confused with it. To construct these pairs, we extract named entities from the test sets using the spaCy \texttt{en\_core\_web\_trf} model~\cite{Honnibal_spaCy_Industrial-strength_Natural_2020}, targeting categories likely to appear in sensitive professional contexts: persons, organisations, locations, products, and events. Each entity (the \textit{acoustic word}) is matched against the CMU Pronouncing Dictionary\footnote{\href{http://www.speech.cs.cmu.edu/cgi-bin/cmudict}{http://www.speech.cs.cmu.edu/cgi-bin/cmudict}}  using Levenshtein distance over ARPAbet phoneme sequences. Substitutes within distance 1 or 2, corresponding to a single phoneme insertion, deletion, or substitution, become the \textit{context word}. To keep search tractable, candidates must share the same first phoneme, and morphological variants (via Porter stemming) are excluded.  Applying this to our three datasets yields 154 pairs from FLEURS (134 at distance 1, 20 at distance 2), 24 from ACL6060 (20/4), and 501 from VoxPopuli (450/51). The context words, 84\% of which are proper nouns, are not inherently sensitive, but sensitivity is domain- and user-dependent: any term in a shared context, from a project name to a client id, may be confidential. Our pairs therefore serve as a controlled evaluation of the phonetic confusion mechanism. 

\begin{table}[t]
\centering
\small
\begin{tabular}{lr}
\toprule
\textit{Evaluation pairs} & 679 pairs \\
\quad Phoneme edit distance 1 / 2        & 604 / 75 \\
\quad Audio (total / avg.\ length)       & 2h 18m / 11.9s \\
\midrule
\textit{Generated context sentences} & \\
\quad Context / acoustic word sentences  & 679 / 679 \\
\quad Filler sentences                   & 6,111 \\
\midrule
\textit{Synthesised audio} & \\
\quad Context / acoustic word recordings & 679 / 679 \\
\midrule
\textit{Prompt-adapted FT data} & \\
\quad Utterances with helpful context    & 1,128 \\
\bottomrule
\end{tabular}
\caption{Statistics of our challenge set for context-induced transcription leakage, comprising 
word pairs from FLEURS ($23$\%), VoxPopuli ($74$\%), and ACL6060 ($3$\%). 
}
\label{tab:dataset_stats}
\end{table}
\subsection{Customisation Approaches}
\paragraph{Context through prompt.}
The model receives background text in the prompt alongside the audio, simulating a user who provides a project brief or a terminology list. We vary the amount of injected context: no context, the context word, a single sentence, five, and ten sentences.
For the context sentence, we use \hfmodel{google/gemma-3-12B-IT}~\cite{gemmateam2025gemma3technicalreport} to generate a thematically relevant sentence that contains the context word, which is verified by string match.  For the 5- and 10-sentence settings, this sentence is inserted at a random position among topically relevant filler sentences, also generated by Gemma with an explicit instruction to exclude both words.

\begin{figure*}[t]
    \centering
    \includegraphics[width=0.9\linewidth]{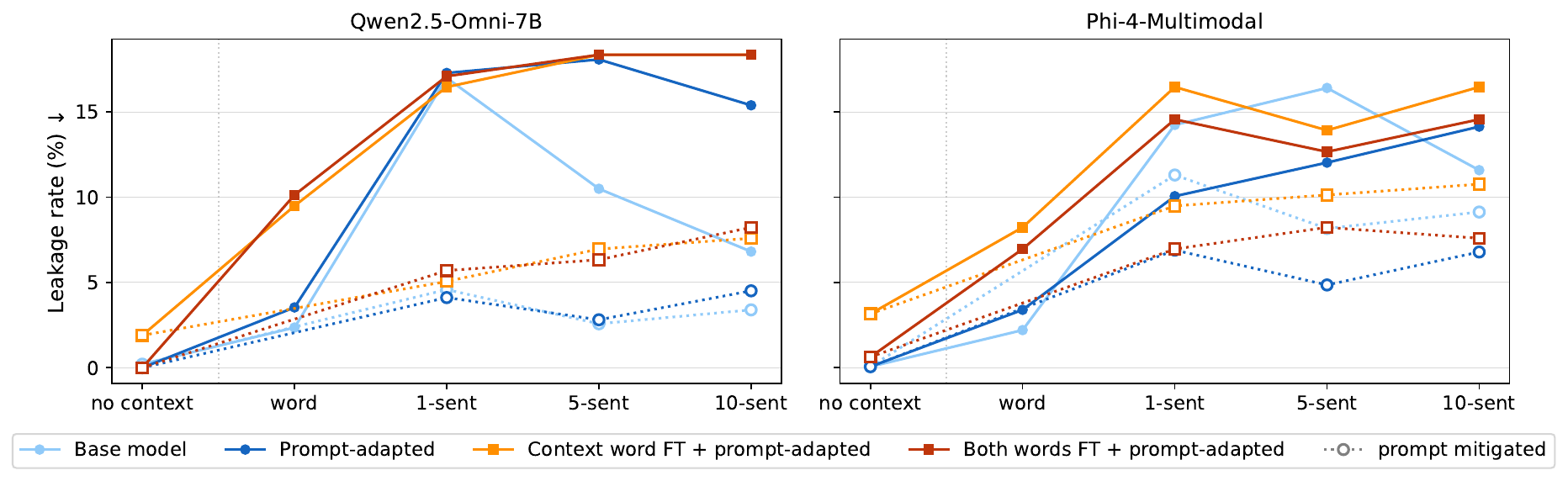}
    \caption{Leakage rate ($\downarrow$) under attack conditions. Context word in prompt and/or fine-tuning on the context.
    }
    \label{fig:mitigation_both}
\end{figure*}

\paragraph{Context through fine-tuning.}
The model is fine-tuned on ASR data in which the context word is spoken, simulating a user who adapts the model to a domain where the term is common. The context prompt sentences from above are reused and corresponding audio is synthesised using \hfmodel{hexgrad/Kokoro-82M} TTS to generate the ASR data.

\paragraph{Combined.}
We evaluate the combination of both customisation mechanisms simultaneously.

\paragraph{Fine-tuning for prompt-following.}
In initial experiments, and consistent with prior work~\cite{gong24b_interspeech, kong2025contextualbiasingllmbasedasr}, we find that SpeechLLMs show an increase in WER when provided with a context prompt (see \cref{sec:baseline}). We therefore fine-tune the models on data where context is helpful (\textit{prompt-adapted model}), applying the NER and context generation pipeline to the FLEURS train split (1,128 utterances).  We generate context passages of 1, 5, and 10 sentences and train on a mixture of all three lengths. Each training example pairs the original audio with its context passage and the correct transcript as the target. The same prompt-adapted model is then used for all three test sets.

Dataset statistics are shown in \cref{tab:dataset_stats}.

\subsection{Models}
We evaluate two recent publicly available SpeechLLMs: \hfmodel{Qwen/Qwen2.5-Omni-7B}~\cite{xu2025qwen25omnitechnicalreport} and \hfmodel{microsoft/Phi-4-multimodal-instruct}~\cite{microsoft2025phi4minitechnicalreportcompact}. We fine-tune both with LoRA~\cite{hu2021loralowrankadaptationlarge}. Fine-tuning and inference are performed on a single NVIDIA A100-SXM4-40GB GPU. Both models are trained 
for two epochs with a learning rate of $1\times10^{-4}$ for Qwen and $4\times10^{-5}$ for Phi.

\begin{figure}[t]
    \centering
    \includegraphics[width=0.9\linewidth]{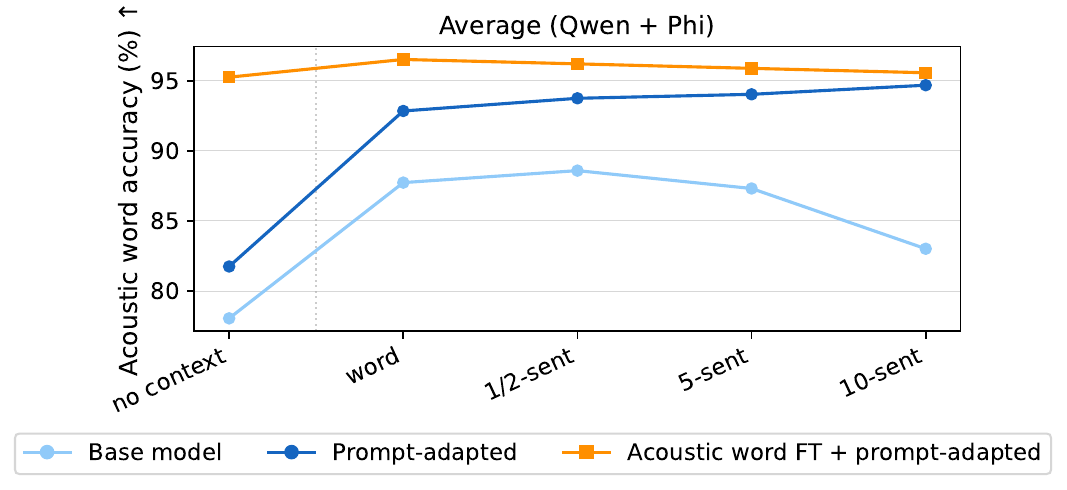}
    \caption{Acoustic word accuracy scores ($\uparrow$) confirm the model can leverage context to improve transcriptions.}
    \label{fig:baseline}
\end{figure}

\subsection{Evaluation}\label{sec:metrics}
To measure general transcription quality independent of leakage, we mask the acoustic word in the reference and both the acoustic and context words in the hypothesis before computing the \textbf{Background WER}~\cite{jiwer}, ensuring that leakage does not influence the WER.

We measure how well the model transcribes acoustic words and how often it leaks context words in its place. For each sample, we find all acoustic word positions $\mathcal{A}$ in the reference using word-level alignment and check whether the hypothesis token $\hat{w}$ matches the acoustic word $w^a$ or the context word $w^c$ to calculate \textbf{acoustic accuracy and leakage rate}:

\begin{equation*}
    \text{Acoustic accuracy} = \frac{|\{i \in \mathcal{A} : \hat{w}_i = w^a_i\}|}{|\mathcal{A}|}
\end{equation*}
\begin{equation*}
    \text{Leakage rate} = \frac{|\{i \in \mathcal{A} : \hat{w}_i = w^c_i\}|}{|\mathcal{A}|}
\end{equation*}

\section{Results}\label{sec:results}

\subsection{Context Improves Acoustic Accuracy}
\label{sec:baseline}

\begin{figure}
    \centering
    \includegraphics[width=1.0\linewidth]{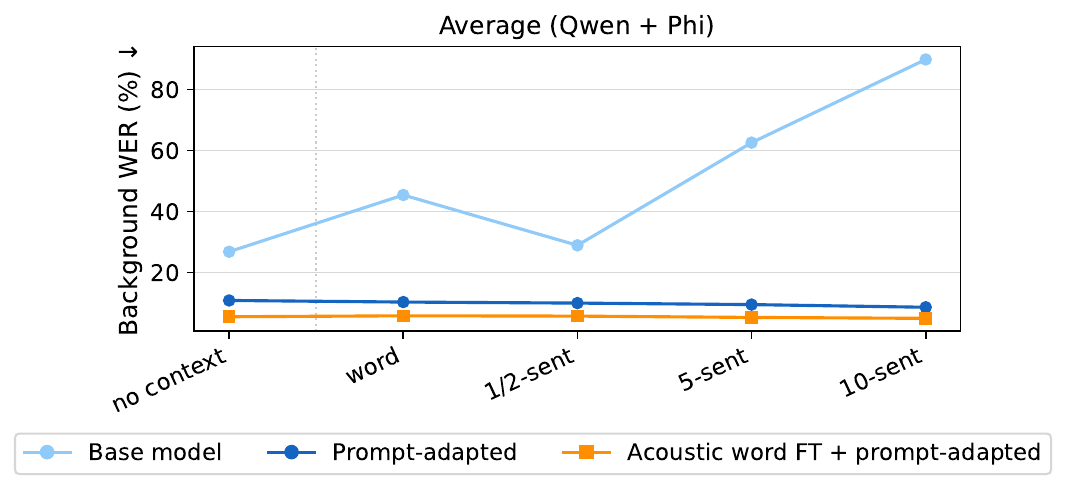}
    \caption{Background WER ($\downarrow$) for the base model and the prompt-adapted models.}
    \label{fig:background_wer}
\end{figure}

We first examine whether providing the acoustic word as context improves ASR, the intended use case for domain customisation, where users supply context to help the model recognise specialised terminology. Acoustic accuracy is shown in \cref{fig:baseline}. Background WER is shown in \cref{fig:background_wer}. Both models show increasing background WER when context is added, confirming prior work \cite{gong24b_interspeech, kong2025contextualbiasingllmbasedasr}. After prompt-adaptation fine-tuning, background WER stabilises at around 9\% across all context lengths. Without context, acoustic word accuracy is already high, and context further improves this. Most of the gain is achieved with a single word or sentence. Additional fine-tuning on audio containing the acoustic word yields a further boost, already in the no-context condition. This confirms that any degradation observed in subsequent experiments is attributable to leakage rather than general transcription difficulty.

\subsection{Privacy Leakage under Context Injection}
\label{sec:leakage}

We now turn to the attack scenario: we inject the context word into the prompt while the acoustic word is spoken, and measure how often the model transcribes the context word instead. Results are shown in \cref{fig:mitigation_both}.

\paragraph{Models are susceptible to prompt-induced leakage.}
The prompt-adapted model shows no leakage without any injected context. Once the context word is introduced, leakage rises and increases further when the word is embedded in a sentence, suggesting that a plausible surrounding sentence makes it more influential. Leakage remains broadly stable at 5 and 10 sentences. The base model shows a qualitatively similar trend but is accompanied by strongly increasing background WER.

\paragraph{Combined customisation amplifies leakage.}
Combining context fine-tuning and context through prompt
substantially amplifies leakage across all conditions, confirming that the two customisation mechanisms compound each other.

\begin{figure}
    \centering
    \includegraphics[width=0.9\linewidth]{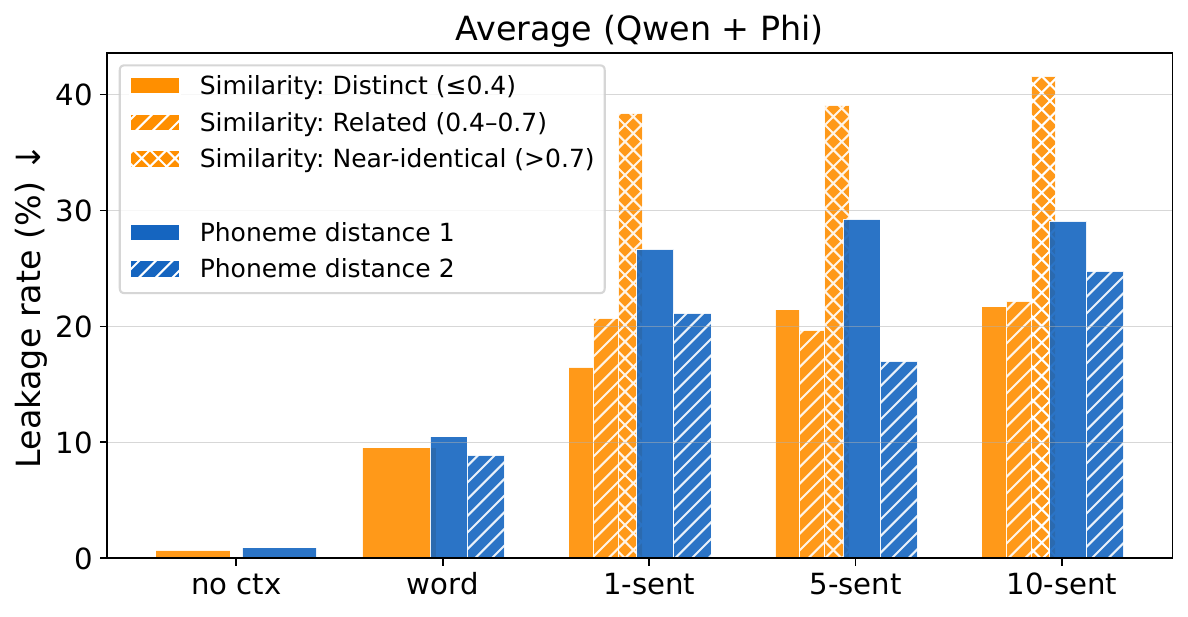}
\caption{Leakage rate  ($\downarrow$) stratified by lexical similarity and phoneme distance between the context and the utterance.}
    \label{fig:similarity_distance_analysis}
\end{figure}

\paragraph{Effect of context sentence similarity and phoneme distance.}
We stratify results along two axes. First, by the lexical similarity between the context sentence and the reference transcript (similarity distribution in \cref{tab:dataset_similarity}). For the combined fine-tuned model (\cref{fig:similarity_distance_analysis}), near-identical context sentences produce substantially higher leakage, indicating that data fine-tuning makes the model more sensitive to how closely the context resembles the utterance. Similar trends hold for the prompt-adapted model.

\begin{table}[h]
\centering
\small
\begin{tabular}{ccc}
\toprule
Distinct ($\leq$0.4) & Related (0.4--0.7) & Similar ($>$0.7) \\
\midrule
333 (49\%) & 160 (24\%) & 186 (27\%) \\
\bottomrule
\end{tabular}
\caption{Lexical similarity between context sentences and spoken reference (char-level overlap) in our test set.}
\label{tab:dataset_similarity}\vspace{-0.2cm}
\end{table}

Second, we stratify by phoneme edit distance between the acoustic and context word (\cref{fig:similarity_distance_analysis}). Leakage is slightly higher for distance-1 pairs than distance-2 pairs, particularly for the combined fine-tuned model, though distance-2 pairs still exhibit substantial leakage. Note that distance-2 contains only 75 pairs versus 604 for distance-1.

\subsection{Accuracy-leakage Trade-off}
\label{sec:mitigation}

We evaluate a potential mitigation strategy to prevent this leakage: including the acoustic word alongside the context word in the prompt. Results are shown in \cref{fig:mitigation_both}.

\paragraph{Adding the acoustic word to the prompt strongly reduces leakage.}
For the prompt-adapted model, adding the acoustic word substantially reduces leakage across all context lengths, as the model now receives competing signals and mostly defaults to the correct transcription. For the combined model, mitigation remains effective, but less so.

\begin{figure}[t]
    \centering
    \includegraphics[width=0.9\linewidth]{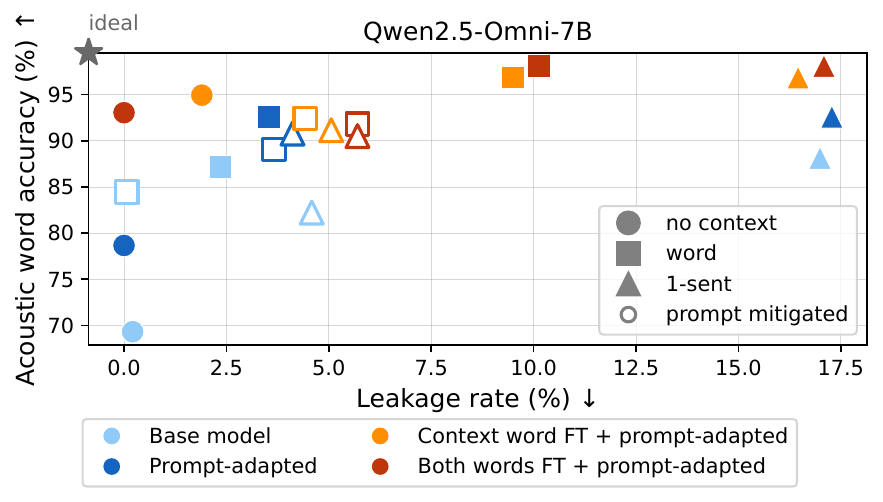}
    \caption{Leakage  ($\downarrow$) vs. accuracy  ($\uparrow$) for different methods.$^3$}
    \label{fig:acc_vs_leakage}
\end{figure}

\paragraph{Fine-tuning provides the best accuracy-leakage trade-off.}
The mitigation strategy comes at a cost: providing both words in the prompt slightly reduces acoustic accuracy compared to providing the acoustic word alone.
\cref{fig:acc_vs_leakage} illustrates the accuracy-leakage trade-off across customisation strategies.\footnote{Leakage and accuracy are measured with different context on the same audio (leakage under attack conditions, accuracy under helpful context) and therefore do not sum to 100\%.} Without prompt injection, domain fine-tuning achieves high acoustic word accuracy with near-zero leakage, making it an effective and safe customisation strategy. 
The privacy risk only materialises when a phonetically similar word is present in the prompt context, whether injected deliberately or introduced incidentally 
through meeting notes or terminology lists.

Additional results stratified by dataset and models can be found in our \href{https://github.com/MaikeZuefle/asr-context-induced-leakage}{repository}.
\section{Conclusion}
Domain customisation is a core feature of modern SpeechLLMs. Our work shows it 
introduces a concrete and previously unexamined privacy risk: the
mechanisms that make a model useful in a specialist domain, context-aware prompting and domain-specific fine-tuning, can cause it to transcribe words from its context or training data rather than what was spoken. 
Our findings suggest that fine-tuning without prompt injection is most effective, but this is unrealistic in practice: context in SpeechLLMs is often used for other tasks than ASR, and fine-tuning is expensive and requires retraining for new terms.
We hope the evaluation framework, our benchmark construction technique, and the resulting dataset provide a foundation for evaluating future approaches.

\section{Acknowledgments}
This work has received funding from the European Union’s Horizon research and innovation programme under grant agreement No 101135798, project Meetween (My Personal AI Mediator for Virtual MEETtings BetWEEN People). 

Generative AI tools were used to assist with editing and grammar checking of the manuscript, as well as for coding and plotting. All scientific content, analyses, and conclusions were developed and verified by the authors.
\bibliographystyle{IEEEtran}
\bibliography{strings,refs}

\end{document}